\documentclass{article} 
\usepackage{nips15submit_e,times}
\usepackage{hyperref}
\usepackage{url}
\usepackage{authblk}
\usepackage{amsmath,graphicx,booktabs}
\usepackage{epstopdf,amssymb}
\usepackage{multirow}
\usepackage{fixltx2e}
\usepackage{subfigure}
\usepackage{float}
\usepackage{titlesec}
\usepackage{bm}

\title{\textbf{A Multi-channel Network with Image Retrieval for Accurate Brain Tissue Segmentation}} 
\author{Yao Sun$^{1,2}$, Yang Deng$^{1,2}$, Yue Xu$^{1,2}$, Shuo Zhang$^{2}$, Mingwang Zhu$^{3}$, Kehong Yuan$^{1*}$\\
	$^{1}$Graduate School at Shenzhen, Tsinghua University, Shenzhen 518055, China.\\
	$^{2}$Department of Biomedical Engineering, Tsinghua University, Beijing 100084, China.\\
	$^{3}$Beijing Sanbo Brain Hospital, Beijing 100825, China.\\
	*Corresponding author: Kehong Yuan (e-mail: yuankh@sz.tsinghua.edu.cn)
}

\nipsfinalcopy 

\begin{document}

\maketitle

\begin{abstract}
Magnetic Resonance Imaging (MRI) is widely used in the pathological and functional studies of the brain, such as epilepsy, tumor diagnosis, etc. Automated accurate brain tissue segmentation like cerebro-spinal fluid (CSF), gray matter (GM), white matter (WM) is the basis of these studies and many researchers are seeking it to the best. Based on the truth that multi-channel segmentation network with its own ground truth achieves up to average dice ratio 0.98, we propose a novel method that we add a fourth channel with the ground truth of the most similar image’s obtained by CBIR from the database. The results show that the method improves the segmentation performance, as measured by average dice ratio, by approximately 0.01 in the MRBrainS18 database. In addition, our method is concise and robust, which can be used to any network architecture that needs not be modified a lot.\\

\end{abstract}

\section{INTRODUCTION}
MRI is an important modern medical imaging technology. Compared with other medical imaging technologies, it has the advantages of high resolution for soft tissue imaging, no radiation damage to human body, and multi-directionality imaging. Because of these advantages, MRI is widely used for pathological and functional studies of the brain, such as research on the working mechanism of the brain and neurological diseases including epilepsy, Alzheimer's disease and multiple sclerosis. The accurate segmentation of MR brain tissue is the basis for these research, diagnosis and treatment. But artificial segmentation is time-consuming and depends on experienced doctors, which is a pain point in clinic. Therefore, an accurate MRI brain tissue automatic segmentation is very important for the auxiliary diagnosis.\\
 	
MRI brain tissues have some special features compared with natural images. Firstly, different soft tissue may have the similar gray information, e.g. the gray information between cerebellum and WM is similar. Secondly, the area of different tissues varies greatly. Thirdly, different tissues’ ground truth is segmented on different modality, e.g. WM lesions are segmented on the FLAIR scan, GM are segmented on the T1 scan. Lastly, the structure of the human brain tissue is relatively fixed, so the texture information between different images has similarity.\\	

Because of the importance of brain tissues segmentation, lots of methods have been developed to achieve accurate segmentation performance in the literature. Broadly speaking, they can be categorized into two classes: 1) Conventional method with hand-crafted features. These methods usually use hand-crafted parameters and get poor results, such as intensity based thresholding \cite{Despotovi2015MRI}, fuzzy c-means clustering \cite{Q2015Automated}, support vector machine (SVM) with spatial and intensity features \cite{Moeskops2015Evaluation}. These methods suffer from limited representation capability for accurate recognition. 2) Deep learning methods with automatically learned features. These methods are based on 2D \cite{Nie2015FULLY} or 3D \cite{Chen2017VoxResNet} CNN to achieve segmentation. Some of these methods have reported state-of-the-art performance on MICCAI MRBrainS13 challenge database \cite{Chen2017VoxResNet,Andermatt2017Multi}, with 1 modality (T1) or 3 modalities (T1, T2-FLAIR, T1-IR). However, these methods don’t exploit the priori fixed brain texture information.\\	

To fully utilize the priori fixed brain texture information, we propose a novel method, which combines CBIR \cite{Zou2009CBIR,Yuan2010Brain,Lin2015Deep,Liu2016Deep} and a multi-channel segmentation CNN network architecture to improve the performance and achieve better segmentation results. We match each slice with the ground truth of its most similar image in the database as the fourth channel (the other 3 channels consist of T1, T2-FLAIR, T1-IR images）. Our approach is concise and effective which can be integrated into any existing network architecture without changing a lot. We segmented 5 tissues including CSF, GM, WM, brain stem and cerebellum, but only the first three categories are evaluated.

\section{METHOD}
Our proposed method consists of basic experiment, achieving similar image with CBIR and multi-channel segmentation network architecture.\\

\textbf{Basic Experiment.} We conducted a basic experiment, in which we trained the multi-channel segmentation network with input of 4 channels including T1, T2-FLAIR, T1-IR and a fourth channel that is its own ground truth (\textbf{Figure 1}) and the average dice ratio is higher than 0.97. It could conclude that the fourth channel with the ground truth provides a priori texture information to the network, making the network performs very well. However, the test data don’t have ground truth, so we proposed a new method which does not require its ground truth and can achieve similar effects. We use CBIR to match each slice with the best similar image’s ground truth. Then the network can use the information provided to facilitate segmentation. 
\begin{figure}[H]
	\begin{center}
		\includegraphics[width=1.\linewidth]{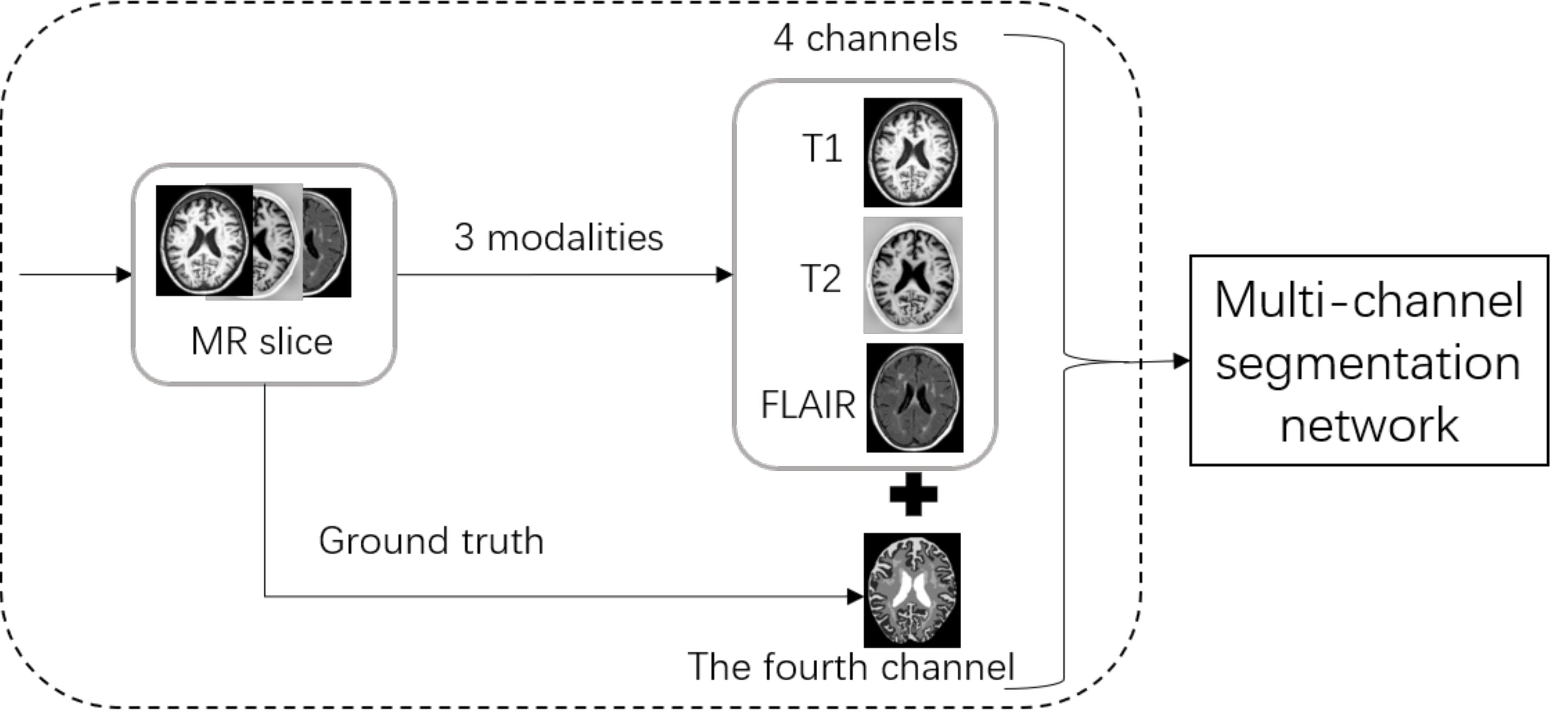}
		\caption{Basic experiment. The input of the forth channel is its ground truth.}
	\end{center}
\end{figure}

\textbf{Achieving similar image with CBIR.} We use CBIR to achieve the best similar image and use its ground truth as the fourth channel of the input. Image retrieval based on CBIR technology: Firstly, the features of the image in the database are extracted, and then the features of the query image are matched with the features of the images in the database, and finally the matched images are obtained. Through CBIR, we solve the problem of needing ground truth, which is concise to achieve. To do this, we first use CBIR to retrieve the best similar image to the test data, and then we register the two images and calculate the similarity of the images after registration. If the similarity is greater than a certain threshold, we will use its ground truth as the fourth channel of the network input (\textbf{Figure 2}), otherwise we will use the original three channels network. 

\begin{figure}[H]
	\begin{center}
		\includegraphics[width=1.\linewidth]{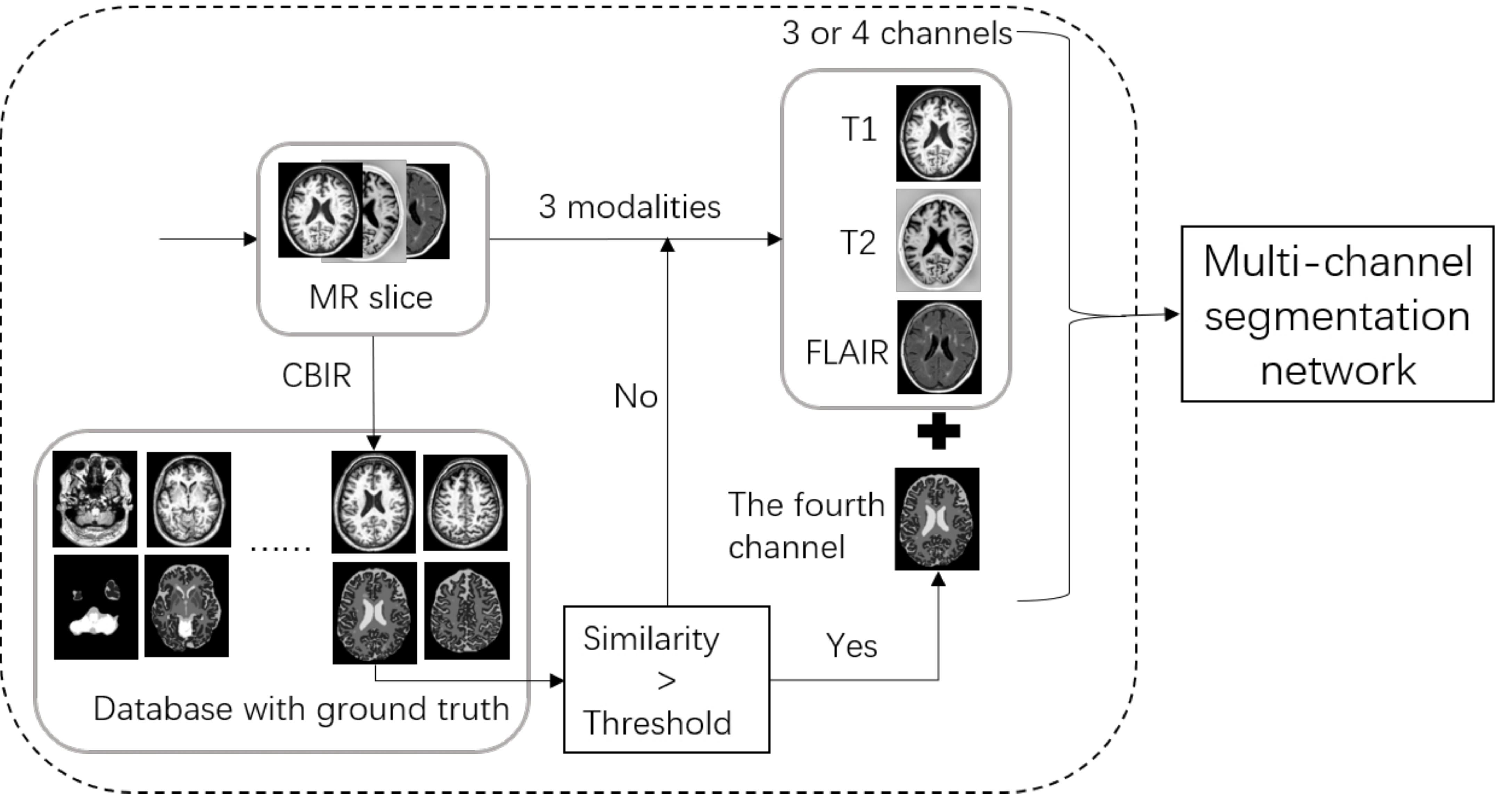}
		\caption{We use CBIR to match the most similar image in the database, if the obtained image’s similarity is larger than the threshold then its ground truth is used as the input of the fourth channel, otherwise we will use the original three channels network.}
	\end{center}
\end{figure}
\textbf{Multi-channel segmentation network.} We designed a multi-channel network (\textbf{Figure 3}) based on U-Net network architecture \cite{Ronneberger2015U} which includes max pooling layers in down-sampling path and deconvolution layers \cite{Zeiler2010Deconvolutional} in up-sampling path. Feature maps of the same resolution from down-sampling and up-sampling layers are concatenated in the up-sampling path to incorporate both coarser and finer information. All the convolution layers are followed by a batch normalization (BN) \cite{Ioffe2015Batch} layer and a ReLu \cite{Nair2010Rectified}. MSE loss function in \textbf{Eqn. (1)} and Adam \cite{Kingma2014Adam} were used during the training.\\

Eqn. (1) is rewritten as:
\begin{equation}
L(x,y)=\frac{1}{N}\sum_{n}^{N}\left \|x _{n}-y_{n} \right \|_{2}^{2}
\end{equation}
where x is the output of the network, y is the target, N is the batch size, L is the loss, and $\left \| \cdot  \right \|_{2}$ is the L2-norm.

\begin{figure}[H]
	\begin{center}
		\includegraphics[width=1.\linewidth]{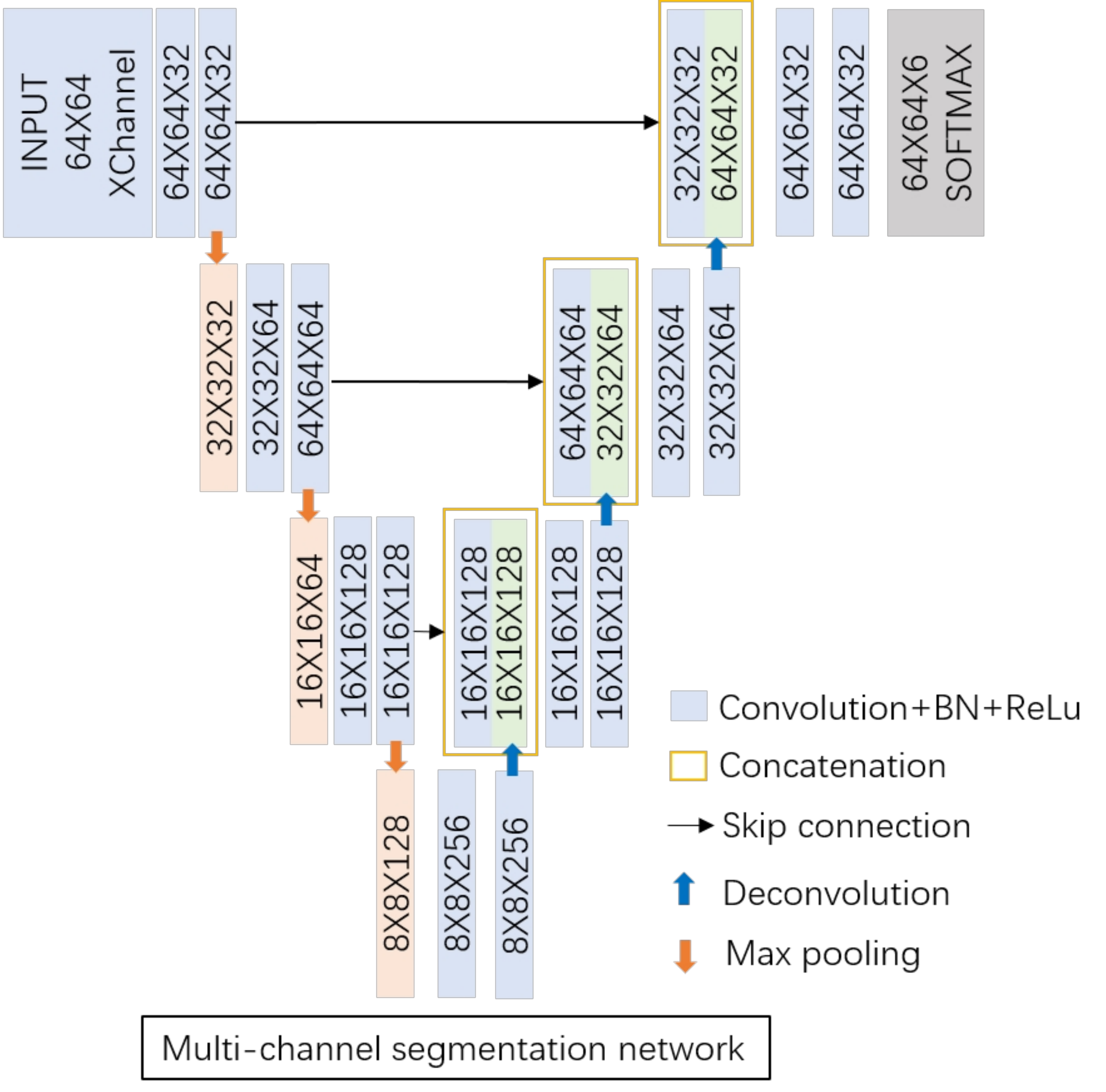}
		\caption{Multi-channel segmentation network. }
	\end{center}
\end{figure}

\section{EXPERIMENTS AND RESULTS}
\subsection{Database Description}
We have evaluated the proposed method on the database of the MICCAI challenge of MR Brain Image Segmentation\footnote{\url{http://mrbrains18.isi.uu.nl/}}. The database provides 5 manual segmentation structures, i.e., background, CSF, GM, WM, brain stem, cerebellum. Multi-sequence 3T MRI brain scans, including T1, T1-IR and T2-FLAIR, are provided for each subject. The train data includes 7 subjects and the test data is unavailable. We split the train data into two groups, 5 (4, 5, 7, 14, 070) for train and 2 (1, 148) for test.\\

\subsection{Implementation Details}
The network is trained with input patch size 64X64XC. C is the channel which is 3 (T1, T2-FLAIR, T1-IR) or 4 (T1, T2-FLAIR, T1-IR, the fourth channel). We use a GPU NVIDIA GeForce TITAN X to train our network. Dice similarity coefficient (DSC) was used as an evaluation metric for the three tissue classes.
Eqn. (1) is rewritten as:

\begin{equation}
DSC=\frac{2TP}{2TP+FP+FN}
\end{equation}

where TP, FP, and FN represent the true positives, false positives, and false negatives of the class for which the score is calculated.
The calculation formula of image similarity is as follows:

\begin{equation}
S(A,B)=1-\frac{\left \| A-B \right \|_{1}}{\left \| A+B \right \|_{1}}
\end{equation}

where A and B are two images, S is the similarity, and $\left \| \cdot  \right \|_{1}$ is the L1-norm. If the two images are completely similar, then S=1. If the two images are completely dissimilar and do not overlap, then S=0.\\

\subsection{Results and Discussion}
Each experiment was performed 5 times and then averaged to guarantee the relia-bility of the proposed method. In order to prove the rationality of our method, we first carried out a basic experiment (\textbf{Table 1}), which used 4 channels including T1, T1-IR, T2-FLAIR and the fourth channel. It should be noted that the fourth channel is its own ground truth. And the dice ratio of CSF, GM and WM over the test data are 0.9874$\pm $0.0046, 0.9747$\pm $0.0564 and 0.9626$\pm $0.0738, respectively, which is robust. We also did an experiment with 3 modalities (T1, T1-IR, T2-FLAIR), and the results were not that good (\textbf{Table 1}). These two experiments show that if we can give the network a fourth channel closest to the inputs’ ground truth, the results should be good. We also did three other experiments, T1, T1-IR, T2-FLAIR, respectively, with their ground truth and the average dice ratios of these three experiments were greater than 0.96. So, we did a third experiment that the fourth channel was a matched image’s ground truth.\\

We used CBIR to match the similar images. We intended to use two methods, \cite{Zou2009CBIR} that matches only similar images and \cite{Lin2015Deep} that matches the best similar image, to provide the fourth channel. We have used the first method \cite{Zou2009CBIR} for the experiment and the image found may not be the most similar. Because the relatively mismatched image, we chose 5 best results in 3 and 4 channels of total 10 experiments as the final results and the average dice ratio is shown in \textbf{Table 1} and the boxplot is shown in \textbf{Figure 4}. Comparing with three modalities experiment, we can see that the dice ratio of our method is improved by adding a ground truth of one of the matched the images to the fourth channel. However, compared with the basic experiment, it still has great potentials for making further progress.\\

So, we assert that if we can match the most similar image, then the results should be better, which we will do in our future work.\\

\begin{figure}[H]
	\begin{center}
		\includegraphics[width=1.\linewidth]{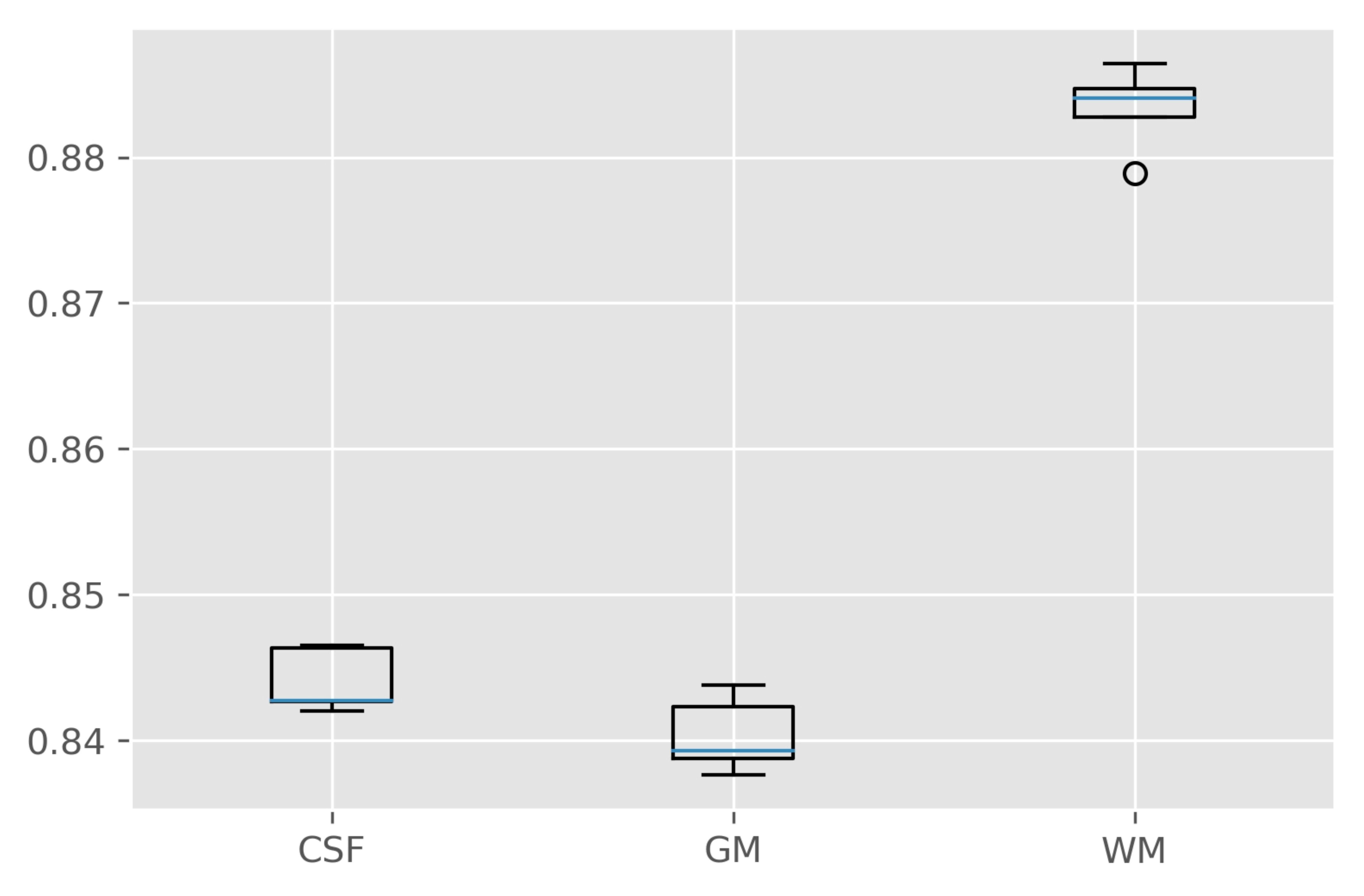}
		\caption{Box plot of 5 best results for CSF, GM and WM dice ratios. CSF: 0.8440$\pm $0.002; GM: 0.8404$\pm $0.0023; WM: 0.8834$\pm $0.0025.}
	\end{center}
\end{figure}

\begin{table}[H]
	\begin{tabular}{ p{0.25\textwidth}  p{0.25\textwidth}  p{0.25\textwidth}  p{0.25\textwidth} } 
		\toprule
		Experiment& CSF & GM & WM\\
		\midrule
		Basic& 0.9874& 0.9747& 0.9626\\
		Three modalities \cite{Chen2017VoxResNet}& 0.8383& 0.8379& 0.8815\\
		Ours& \textbf{0.8440}& \textbf{0.8404}& \textbf{0.8834}\\
		\bottomrule
	\end{tabular}
	\caption{Dice ratio comparison of our method with basic and three modalities experiment. Basic: with its ground truth as the fourth channel; Three modalities: with traditional three channels (T1, T1-IR, T2-FLAIR); Ours: with similar image’s ground truth as the fourth ground truth. }
\end{table}

By visualizing the results, we observe that the boundaries between different tissues are smoother and the random noise in the continuous tissues has been solved by adding the fourth channel, which proves that the fourth channel can provide valid prior texture information of which the network can take advantage (\textbf{Figure 5}). But we find that some tissues have been misclassified, which, we think, is because of the mismatched texture. And it should be solved by matching the most similar image.
\begin{figure}[H]
	\begin{center}
		\includegraphics[width=1.\linewidth]{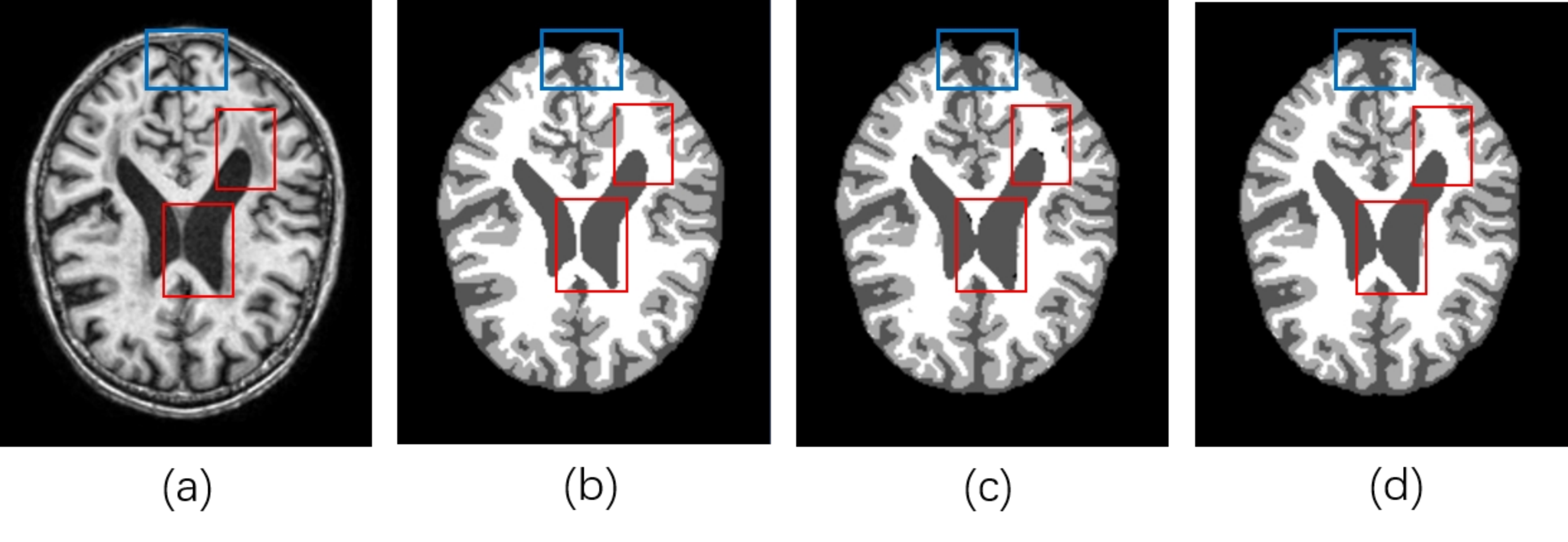}
		\caption{Visualization of (a) Input, (b) Ground Truth; Black: background, Dark gray: CSF, Gray: GM, White: WM, (c) 3 channels, (d) 4 channels. Some difference is shown in rectangles between 3 channels and 4 channels. The additional forth label channel helps the segmentation network to capture smooth texture information and solve the random noise problem in the boundary between two tissues. But in the blue rectangles, some of the skull part was mistakenly divided into CSF. }
	\end{center}
\end{figure}

\section{CONCLUSION}
In this paper, we proposed a novel method that is helpful in accurate brain tissue segmentation. The fourth channel can provide prior texture information to promote smooth segmentation of different tissues boundaries. However, some tissues are misclassified due to the mismatched information. We only match the slices using general CBIR method, but the results ware increased by 0.01.\label{key} We believe that the results will be better if the fourth channel is the most similar images’ ground truth, which is what we plan to do in our future work. Our method is concise and can be implemented to almost any network architecture without changing these models a lot.

\bibliographystyle{IEEEtranS}
\bibliography{3D-DRL}

\end{document}